\documentclass{article}

\usepackage{microtype}
\usepackage{graphicx}
\usepackage{subcaption}
\usepackage{booktabs}
\usepackage{multirow}
\usepackage{tabularx}
\usepackage{array}
\usepackage{xcolor}
\usepackage{colortbl}
\usepackage{enumitem}
\usepackage{listings}
\usepackage{placeins}

\usepackage{hyperref}
\usepackage{xurl}

\usepackage[preprint]{icml2026}

\usepackage{amsmath}
\usepackage{amssymb}
\usepackage{mathtools}
\usepackage{amsthm}
\usepackage{bm}

\definecolor{jamsblue}{HTML}{0055B3}
\definecolor{jamsrow}{HTML}{EEF5FE}
\definecolor{jamstint}{HTML}{EBF3FC}
\definecolor{jamswarn}{HTML}{FCE8EE}

\hypersetup{
  colorlinks=true,
  linkcolor=jamsblue,
  citecolor=jamsblue,
  urlcolor=jamsblue,
  pdftitle={JAMS: Optimizing Training-Free Decision Models with Judge-Gated Agentic Middleware Search},
  pdfauthor={Tu Hu, Zhen Zhao, Linjian Meng, Yuqi Fang},
  pdfsubject={Preprint}
}

\usepackage[capitalize,noabbrev]{cleveref}

\newcommand{\up}{\ensuremath{\uparrow}}

\icmltitlerunning{JAMS: Optimizing Training-Free Decision Models}

\begin{document}

\twocolumn[
  \icmltitle{JAMS: Optimizing Training-Free Decision Models \\
    with Judge-Gated Agentic Middleware Search}

  \begin{icmlauthorlist}
    \icmlauthor{Tu Hu}{nju,sail}
    \icmlauthor{Zhen Zhao}{sail}
    \icmlauthor{Linjian Meng}{sail}
    \icmlauthor{Yuqi Fang}{nju}
  \end{icmlauthorlist}

  \icmlaffiliation{nju}{Nanjing University}
  \icmlaffiliation{sail}{Shanghai AI Laboratory}
  \icmlcorrespondingauthor{Linjian Meng}{menglinjian@pjlab.org.cn}
  \icmlcorrespondingauthor{Yuqi Fang}{yqfang@nju.edu.cn}

  \icmlkeywords{Decision Models, Training-Free Methods, Language Models, Agentic Search, Program Optimization}

  \vskip 0.3in
]

\printAffiliationsAndNotice{}

\begin{abstract}
Decision models map an input and a predefined set of options to a probability distribution over those options; they power classification, routing, scoring, and verification in software systems, and the recent release of Jev has drawn wide attention to this model class. Training-free methods build such models directly from off-the-shelf language models, so that practitioners can use increasingly capable base models immediately, without collecting data or training. Existing training-free methods, however, only adapt the base model to the decision format through a fixed, hand-designed procedure and leave the procedure itself unoptimized, although this choice alone substantially changes decision quality and cost. We propose Judge-gated Agentic Middleware Search (JAMS), which treats the decision procedure as the object of optimization: a coding agent iteratively proposes and refines candidate procedures, an independent judge admits only candidates that genuinely rely on the model rather than on task-specific rules, and the search retains the candidates with the best quality--efficiency trade-offs. A single procedure discovered on one model transfers unchanged to five open models and six unseen tasks, raising the average score from 51.9\% for LLM2Jev to 60.0\% while running 5.3$\times$ faster, and attains the highest average score among all evaluated training-free methods on every model.
\end{abstract}

\section{Introduction}
\label{sec:intro}

Many components of modern software do not need free-form text from a language model; they need a decision. Routing a request to the right tool, flagging an unsafe message, scoring a response against a rubric, and checking whether a claim is supported all reduce to choosing among a predefined set of options. \emph{Decision models} serve exactly this purpose: given an input and a set of candidate options, they return a probability distribution over the options that downstream code can threshold, rank, or aggregate \citep{almeida2026jev,li2026llmasjev}. Because the output space is fixed in advance, their answers are always well formed, their confidence is explicit, and each query is inexpensive. The recent release of Jev \citep{almeida2026jev}, a decision model built for fast, typed probabilistic decisions, has drawn wide attention to this model class and raised a practical question: how can strong decision models be obtained?

\begin{figure*}[t]
  \centering
  \includegraphics[width=\textwidth]{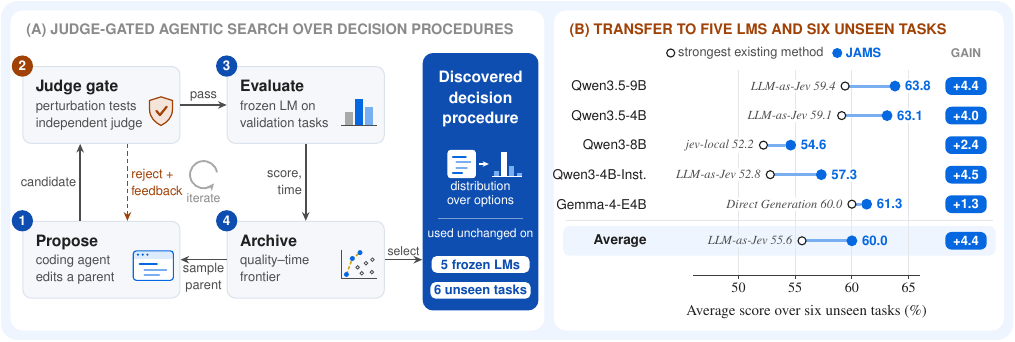}
  \caption{\textbf{JAMS optimizes the decision procedure of a training-free decision model, and the discovered procedure transfers across models.} \textbf{(A)}~Starting from an empty archive, a coding agent proposes candidate procedures; a judge gate rejects candidates that bypass the model or depend on task-specific rules and returns feedback for repair; admitted candidates are evaluated for decision quality and inference time, and the archive keeps the quality--time frontier from which the next parent is sampled. \textbf{(B)}~Average score over six unseen tasks of JAMS and of the strongest of seven existing training-free methods on each model. The strongest existing method differs across models; JAMS, applied unchanged, exceeds it on every model.}
  \label{fig:overview}
\end{figure*}

Building decision models \emph{without training} offers an immediate alternative. General-purpose language models already behave as capable decision models when queried appropriately \citep{li2026llmasjev}, and training-free methods exploit this property: they need no labeled data or gradient updates, apply to any open model, and inherit every improvement of the base model as soon as a stronger one is released. Several such methods have been proposed. Direct Generation asks the model to write the index of its answer \citep{li2026llmasjev}; SemIf \citep{semif2026} and open-alt-jev \citep{openaltjev2026} read the probability of each option label in a single forward pass; LLM-as-Jev scores the full answer string of every option after a fixed answer prefix \citep{li2026llmasjev}; LLM2Jev asks the model a separate yes/no question for each option \citep{yinsongxu2026llm2jev}; jev-local scores the likelihood of each option text \citep{jevlocal2026}; and AnyJev averages over permuted option orders and calibrates against content-free inputs \citep{zhang2026anyjev,zhao2021calibrate}.

Despite their differences, these methods share a common limitation: they \emph{adapt} the base model to the decision format but do not \emph{optimize} how the model is used to reach a decision. Each method fixes a hand-designed decision procedure---how the options are presented, how often the model is queried, and how its outputs are combined---and applies it unchanged to every model and task. This choice matters far more than its fixed status suggests: on each of the five models we study, the average score of existing procedures varies by 13.9 to 34.0 points and their time per question by a factor of 10 to 18, and the best procedure differs across models (\Cref{sec:motivation}). The decision procedure is thus a consequential choice that no method optimizes.

Optimizing the decision procedure, however, poses two challenges. First, the design space is structural rather than textual: a better procedure may query the model several times, reorder or prune the options, or combine evidence across queries. Such changes lie outside the reach of prompt optimizers, which refine instructions under a fixed query structure \citep{zhou2023large,yang2024opro,khattab2024dspy,agrawal2026gepa}. Second, an optimizer that is expressive enough to restructure the procedure is also expressive enough to overfit. In a pilot study (\Cref{sec:motivation}), an unconstrained coding agent learned to answer an entire task with hand-written rules that never consulted the model; these rules scored highly on held-out cases of that task but contributed nothing on unseen tasks. Such solutions defeat the purpose of a decision model, whose value lies in the general capability of the underlying language model. An effective optimizer must therefore search over expressive procedures while ensuring that the model makes every decision.

We propose \textbf{Judge-gated Agentic Middleware Search (JAMS)}, a framework that automatically optimizes the decision procedure of training-free decision models (\Cref{fig:overview}A). Concretely, JAMS represents a decision procedure as a program that sits between the application and the frozen model---a piece of \emph{decision middleware}---and that accesses the model only through a small set of operations for scoring options and generating text. Starting from scratch, JAMS runs an iterative search loop. A coding agent proposes a new procedure by editing a parent drawn from an archive of previous candidates. A judge gate, which combines perturbation tests with an independent reviewer, rejects candidates that bypass the model or rely on task-specific rules and returns diagnostic feedback for repair. Admitted candidates are evaluated on validation tasks for both decision quality and inference time, and the archive retains those that offer the best trade-offs, from which the next parent is sampled. The output of the search is a single procedure that is reused unchanged across models and tasks.

We run the search on two open models and apply the discovered procedures unchanged to five open models from two families and six tasks never seen during search, covering choices among 2 to 151 options, rating scales, and binary verification. JAMS attains the highest average score among seven training-free methods on all five models: it exceeds the strongest existing method of each model by 1.3 to 4.5 points and raises the five-model average from 55.6\% for LLM-as-Jev, the strongest baseline on average, to 60.0\%, while answering 4.0$\times$ faster on average (3.1--5.3$\times$ across models) than LLM-as-Jev. Its gains are largest where many options must be compared: on the 151-option intent task, it raises the score from 71.5\% for LLM2Jev to 80.9\% while cutting model calls per question from 151 to 7. The discovered program combines strategies that no existing method uses together, such as screening many options before re-scoring a short list and averaging over option orders.

\noindent\textbf{Contributions.}\nopagebreak
\begin{itemize}[leftmargin=1.1em, itemsep=2pt, topsep=2pt, parsep=0pt, beginpenalty=10000]
  \item We identify the decision procedure as an underexplored optimization target for training-free decision models: hand-designed procedures leave substantial headroom in quality and efficiency, and unconstrained optimization of the procedure is prone to shortcuts that do not transfer.
  \item We propose JAMS, an agentic search framework that optimizes decision procedures from scratch, with a judge gate that restricts the search to model-driven, task-agnostic procedures and a quality--time archive that guides exploration.
  \item We show that a single procedure discovered by JAMS transfers across models and unseen tasks and attains the highest average score among existing training-free methods on every model, and we analyze the strategies that it discovers.
\end{itemize}

\section{Related Work}
\label{sec:related}

\textbf{Decision models and training-free adaptation.}
Decision models map requests to probability distributions over fixed option sets; Jev popularized this typed interface \citep{almeida2026jev}, and \citet{li2026llmasjev} show that general-purpose language models already behave as Jev-style decision models without fine-tuning. A closely related line of work uses language models as judges that score or compare outputs \citep{zheng2023judging}. Training-free decision methods (\Cref{sec:intro}) differ mainly in how they elicit a distribution from the model: by generating an answer, possibly under constrained decoding \citep{willard2023efficient}, or by reading the probabilities of option labels, option texts, or per-option yes/no answers. These choices interact with well-documented weaknesses of language models, including competition between surface forms \citep{holtzman2021surface}, sensitivity to option order \citep{pezeshkpour2024large,zheng2024large}, and miscalibrated label priors, which contextual calibration and its extensions correct \citep{zhao2021calibrate,zhang2026anyjev}. Each existing method commits to one hand-designed procedure. In contrast, JAMS treats the procedure as a search space and lets the optimizer decide which of these ingredients to use and how to combine them.

\textbf{From prompt optimization to program search.}
A growing body of work optimizes how a frozen language model is used rather than its weights. Prompt optimizers search over instructions and demonstrations with language-model proposals \citep{zhou2023large,yang2024opro}, textual feedback \citep{yuksekgonul2024textgrad}, compiled pipelines \citep{khattab2024dspy}, or reflective evolution \citep{agrawal2026gepa}, and recent methods evolve richer contexts that accumulate strategies over time \citep{zhang2025ace}. These methods typically keep the structure of model calls fixed. Program search goes further by letting a language model write and refine code, which has produced new algorithms \citep{romeraparedes2024mathematical,novikov2025alphaevolve}, agent designs and workflows \citep{hu2025automated,zhang2025aflow}, and task-specific model harnesses \citep{lee2026metaharness}. JAMS applies program search to the decision procedure of training-free decision models, with a different goal from most prior work: rather than solving a target benchmark, it seeks a single procedure that remains effective on models and tasks never seen during search, which requires explicit safeguards against overfitting to the tasks used for optimization.

\textbf{Specification gaming and independent review.}
Optimizing against a measurable proxy invites solutions that improve the proxy without improving the intended behavior \citep{pan2022effects,skalse2022defining}. The risk grows with the expressiveness of the optimizer: a program-level optimizer can encode knowledge about the evaluation tasks directly into code. JAMS addresses this risk with a gate that combines behavioral tests, which check how a candidate responds to controlled changes of its input, with an independent language-model reviewer \citep{zheng2023judging} that inspects the code itself. Only candidates that pass both checks are evaluated and allowed to influence the search.

\section{Problem Formulation and Motivation}
\label{sec:motivation}

\begin{figure*}[t]
  \centering
  \includegraphics[width=\textwidth]{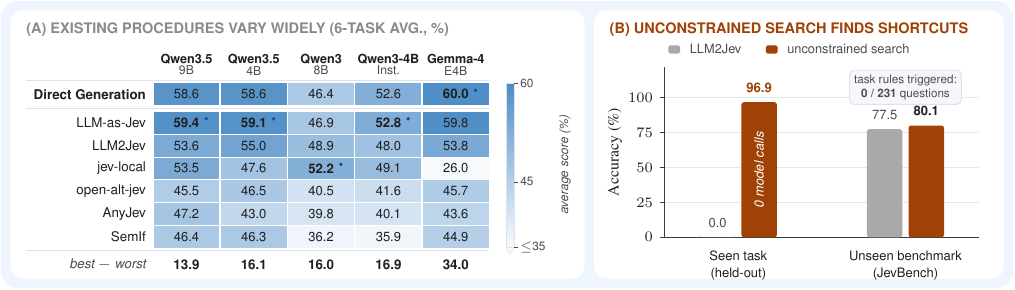}
  \caption{\textbf{The decision procedure matters, but optimizing it without constraints leads to shortcuts.} \textbf{(A)}~Average score over six unseen tasks of seven existing procedures on five models, colored by score; the best procedure of each model is in bold with a dot, and the bottom row gives the gap between the best and worst procedure. \textbf{(B)}~Without constraints, the search wrote a rule-based solver that answers held-out cases of a household-control task with 96.9\% exact match and zero model calls; on an unseen benchmark (JevBench), none of its task rules ever triggered.}
  \label{fig:motivation}
\end{figure*}

\subsection{Training-Free Decision Models}
\label{sec:formulation}

A decision request $x = (c, \mathcal{O})$ consists of a context $c$, which holds the input and an instruction, and a set of options $\mathcal{O} = \{o_1, \dots, o_K\}$. We consider the three kinds of decisions that are common in practice: a \emph{choice} among $K$ options, a \emph{rating} on an ordered scale of $K$ levels that is summarized by its expected value, and a binary \emph{verification} whose two options are ``true'' and ``false''. A decision model answers a request with a probability distribution over the options. A training-free decision model obtains this distribution from a frozen language model $f_\theta$ through a \emph{decision procedure} $M$:
\begin{equation}
  p = M(x; f_\theta) \in \Delta^{K-1}.
  \label{eq:procedure}
\end{equation}
The procedure specifies how the request is presented to the model, how many times and in which form the model is queried, and how the outputs of the model are combined into $p$. Every existing training-free method corresponds to one fixed procedure (\Cref{sec:method_space}).

We evaluate a procedure by its decision quality $Q(M; \mathcal{T}, f_\theta)$, the task metric averaged over a set of tasks $\mathcal{T}$, and by its cost $C(M; \mathcal{T}, f_\theta)$, the average wall-clock time per question when requests are served concurrently. Because decision models are routinely applied to new tasks and, as better open models appear, to new base models, what matters is the quality of a procedure on tasks and models that were not available when it was designed. Our goal is therefore to find, using a set of source tasks and a single search model, a procedure that performs well when applied unchanged to unseen tasks and other models, at a cost no higher than that of existing methods.

\subsection{The Decision Procedure Matters}
\label{sec:motivation_headroom}

We start with a diagnostic question: how strongly does the procedure affect a decision model when the base model is held fixed? \Cref{fig:motivation}A compares the seven existing training-free procedures on five open models and six tasks (setup in \Cref{sec:setup}). Two observations stand out. First, the differences are large: on every model, the average score of the seven procedures varies by 13.9 to 34.0 points (\Cref{tab:main}), and their time per question varies by a factor of 10 to 18 (\Cref{sec:time}). Second, no procedure is best everywhere: the strongest method is LLM-as-Jev on three models, jev-local on one, and Direct Generation on the remaining one. The quality of a training-free decision model thus hinges on a design choice that is currently made by hand, once, and independently of the base model. These differences motivate optimizing the procedure itself rather than choosing among fixed designs.

\subsection{Unconstrained Optimization Finds Shortcuts}
\label{sec:motivation_shortcut}

A natural way to optimize the procedure is to let a coding agent rewrite it and to keep the versions that perform best on held-out data. We tested this idea in a pilot study that placed no constraint on the generated code and used several tasks with separate training and held-out cases (\Cref{sec:app_pilot}). The search quickly produced programs that answer questions without consulting the model. For a household-control task whose inputs follow a fixed structure, the agent wrote a rule-based solver that parses the request and device states directly. This solver reached 96.9\% exact match on held-out cases without a single model call, whereas LLM2Jev solved none of them (\Cref{fig:motivation}B). Because the solver is both more accurate and faster on held-out cases, ordinary validation prefers it. Yet it encodes knowledge about one task rather than a better way of using the model: on an unseen benchmark, none of its task-specific rules ever triggered, and its modest gain there came entirely from its model queries.

The pilot exposes a general tension. An optimizer that is free to restructure the procedure is also free to memorize the structure of its training tasks, and held-out cases drawn from the same tasks cannot reveal this. We therefore restrict the search to procedures that are \emph{model-driven}, meaning that their decisions are determined by the outputs of the language model, and \emph{task-agnostic}, meaning that they do not branch on the identity of a task or rely on rules and vocabularies written for a particular task. \Cref{sec:method} describes how JAMS enforces these requirements while searching for better procedures.

\section{JAMS: Judge-Gated Agentic Middleware Search}
\label{sec:method}

JAMS optimizes the decision procedure $M$ of \Cref{eq:procedure} by searching over programs (\Cref{fig:overview}A; \Cref{alg:jams}). We define the search space and then describe how candidates are proposed, gated, and selected.

\subsection{Search Space: Decision Procedures as Programs}
\label{sec:method_space}

Each candidate is a program that receives a decision request and returns a distribution over its options. The program may perform arbitrary computation on the request, but it can interact with the frozen model only through a small interface of three operations: \emph{scoring}, which returns the probabilities that the model assigns to a given set of tokens as the next token after a prompt, without generating any text; \emph{tokenization}, which maps a string to the tokens of the model; and \emph{generation}, which produces a short continuation. Because the interface exposes the model only through these generic operations, every program can run unchanged on any open model that provides them.

This space contains all existing training-free methods as special cases. Direct Generation is a single generation call; SemIf is a single scoring call over the option labels; LLM2Jev is one scoring call per option. At the same time, the space admits procedures that no fixed method implements: a program may query the model several times with different prompts, discard or reorder options between queries, combine probabilities across queries, or adapt its strategy to the number of options and to how the model tokenizes them. Candidates run in an isolated environment without network or data access and with a bounded number of model calls per request (\Cref{sec:app_impl}).

\subsection{Agentic Proposal}
\label{sec:method_propose}

JAMS maintains an archive $\mathcal{C}$ of admitted candidates, each annotated with its validation quality and inference time. The archive starts empty: the search is not seeded with any existing method, so every strategy in the final procedure is discovered by the search itself. At each iteration, JAMS samples a parent uniformly from the quality--time Pareto frontier of the archive, which keeps both accurate and fast procedures available for further improvement \citep{agrawal2026gepa}. It then draws a mini-batch of training requests from the source tasks, with task names and identifiers removed. A coding agent receives the parent program, the mini-batch together with the decisions, errors, and inference time of the parent on it, and the specification of the model interface. The agent analyzes the failures and writes a new program, which it can execute on the mini-batch before submitting. Because the agent edits code rather than text, a single proposal can restructure the control flow---for instance, adding a screening pass or averaging across option orders.

\begin{algorithm}[t]
  \caption{Judge-Gated Agentic Middleware Search}
  \label{alg:jams}
  \begin{algorithmic}[1]
    \STATE {\bfseries Input:} source tasks with training and validation splits, search model $f_\theta$, iterations $N$, perturbation tests $\mathcal{P}$, judge $J$, reference procedure $M_{\mathrm{ref}}$
    \STATE $\mathcal{C} \leftarrow \emptyset$ \hfill \textit{// no seed procedure}
    \FOR{$n = 1$ {\bfseries to} $N$}
      \STATE $M_{\mathrm{par}} \leftarrow$ sample from the Pareto frontier of $\mathcal{C}$ (none if $\mathcal{C} = \emptyset$)
      \STATE $B \leftarrow$ training requests with task identities removed
      \STATE $\mathit{feedback} \leftarrow \emptyset$
      \REPEAT
        \STATE $M \leftarrow \textsc{Agent}(M_{\mathrm{par}}, B, \mathit{feedback})$
        \STATE $\mathit{ok}, \mathit{feedback} \leftarrow \textsc{Gate}(M; \mathcal{P}, J)$
      \UNTIL{$\mathit{ok}$ or the repair budget is exhausted}
      \IF{$\mathit{ok}$}
        \STATE $(\alpha, \tau) \leftarrow \textsc{Evaluate}(M; f_\theta, \text{validation split})$
        \STATE $\mathcal{C} \leftarrow \mathcal{C} \cup \{(M, \alpha, \tau)\}$
      \ENDIF
    \ENDFOR
    \STATE {\bfseries return} $M^\star$ and $M^\star_{\mathrm{fast}}$ selected from $\mathcal{C}$ by \Cref{eq:selection}
  \end{algorithmic}
\end{algorithm}

\subsection{Judge Gate}
\label{sec:method_gate}

Every candidate must pass a gate before it is evaluated on validation data; candidates that fail are never scored and therefore cannot influence the search. The gate combines two complementary checks.

\textbf{Perturbation tests} examine the behavior of a candidate. The candidate is executed on synthetic requests from fictional tasks, with the language model replaced by a simulated one, and under meaning-preserving perturbations: renamed input fields, paraphrased instructions, and shuffled options. A model-driven and task-agnostic procedure must consult the model before producing any non-uniform decision, and its decisions must follow the options rather than their positions or the names of input fields. The tests record every case in which a candidate breaks these properties.

\textbf{An independent judge} examines the code. A language model from a different family than the coding agent reviews the program source together with the test records, following a fixed rubric of prohibited behaviors: answering without the model, branching on task content or field names, task-specific vocabularies or prompts, memorized cases, surface heuristics such as preferring the longest option, and information carried across requests (the full rubric is given in \Cref{sec:app_gate}). The judge has no access to data, scores, or the agent's reasoning, so its verdict cannot be swayed by how well a candidate performs.

The two checks cover each other's blind spots. Tests observe what a program does but can miss logic that is triggered only by real task inputs, whereas code review can recognize the intent of such logic but may overlook subtle numerical effects. When either check fails, its diagnostics are returned to the agent, which may repair the candidate within the same iteration. Before running the search, we calibrated the gate on the shortcut programs collected in the pilot study: it rejected all of them while admitting generic procedures (\Cref{sec:app_gate}).

\begin{table*}[t]
  \caption{\textbf{JAMS attains the highest average score on every model.} Average score over six unseen tasks (\%); JAMS is discovered once on Qwen3.5-9B and applied unchanged to every model. Calls: model calls per question with $K$ options. Max.\ $K$: largest number of options a method accepts; larger questions (CLINC150+OOS, 151 options) count as incorrect. Prob.: graded probabilities rather than a single answer. Gain: JAMS minus the strongest existing method of each column. Qwen3-4B-Inst.\ and Gemma-4-E4B denote Qwen3-4B-Instruct-2507 and Gemma-4-E4B-it. Bold: best; underlined: second best. Per-task results are given in \Cref{sec:app_per_scenario}.}
  \label{tab:main}
  \centering
  \small
  \setlength{\tabcolsep}{4.4pt}
  \renewcommand{\arraystretch}{1.14}
  \resizebox{\textwidth}{!}{%
  \begin{tabular}{@{}l ccc rrrrr r@{}}
    \toprule
    & \multicolumn{3}{c}{Procedure} & \multicolumn{6}{c}{Average score over six unseen tasks (\% \up)} \\
    \cmidrule(lr){2-4} \cmidrule(l){5-10}
    Method & Calls & Max.\ $K$ & Prob. & Qwen3.5-9B & Qwen3.5-4B & Qwen3-8B & Qwen3-4B-Inst. & Gemma-4-E4B & Avg. \\
    \midrule
    Direct Generation & 1 & any & $\times$ & 58.6 & 58.6 & 46.4 & 52.6 & \underline{60.0} & 55.2 \\
    SemIf~{\scriptsize\citep{semif2026}} & 1 & 16 & \checkmark & 46.4 & 46.3 & 36.2 & 35.9 & 44.9 & 41.9 \\
    open-alt-jev~{\scriptsize\citep{openaltjev2026}} & 1 & 26 & \checkmark & 45.5 & 46.5 & 40.5 & 41.6 & 45.7 & 44.0 \\
    LLM-as-Jev~{\scriptsize\citep{li2026llmasjev}} & $K{+}1$ & any & \checkmark & \underline{59.4} & \underline{59.1} & 46.9 & \underline{52.8} & 59.8 & \underline{55.6} \\
    LLM2Jev~{\scriptsize\citep{yinsongxu2026llm2jev}} & $K$ & any & \checkmark & 53.6 & 55.0 & 48.9 & 48.0 & 53.8 & 51.9 \\
    jev-local~{\scriptsize\citep{jevlocal2026}} & $K$ & any & \checkmark & 53.5 & 47.6 & \underline{52.2} & 49.1 & 26.0 & 45.7 \\
    AnyJev~{\scriptsize\citep{zhang2026anyjev}} & $4K$ & 26 & \checkmark & 47.2 & 43.0 & 39.8 & 40.1 & 43.6 & 42.7 \\
    \midrule
    \rowcolor{jamsrow} JAMS (ours) & 1--7 & any & \checkmark & \textbf{63.8} & \textbf{63.1} & \textbf{54.6} & \textbf{57.3} & \textbf{61.3} & \textbf{60.0} \\
    \rowcolor{jamsrow} \quad {\small\textcolor{jamsblue}{Gain}} & & & & \textcolor{jamsblue}{\textbf{+4.4}} & \textcolor{jamsblue}{\textbf{+4.0}} & \textcolor{jamsblue}{\textbf{+2.4}} & \textcolor{jamsblue}{\textbf{+4.5}} & \textcolor{jamsblue}{\textbf{+1.3}} & \textcolor{jamsblue}{\textbf{+4.4}} \\
    \bottomrule
  \end{tabular}}
\end{table*}

\subsection{Evaluation and Selection}
\label{sec:method_select}

An admitted candidate $M_i$ is evaluated with the search model on validation data from the source tasks, yielding its quality $\alpha_i$ and time per question $\tau_i$, and is added to the archive. After $N$ iterations, JAMS selects two procedures relative to the reference procedure $M_{\mathrm{ref}}$ (LLM2Jev, with quality $\alpha_{\mathrm{ref}}$ and time per question $\tau_{\mathrm{ref}}$). The primary procedure maximizes quality without being slower than $M_{\mathrm{ref}}$, and a fast variant minimizes time per question without being less accurate:
\begin{equation}
  M^\star = \operatorname*{arg\,max}_{i:\, \tau_i \le \tau_{\mathrm{ref}}} \alpha_i, \qquad
  M^\star_{\mathrm{fast}} = \operatorname*{arg\,min}_{i:\, \alpha_i \ge \alpha_{\mathrm{ref}}} \tau_i .
  \label{eq:selection}
\end{equation}
We refer to $M^\star$ as JAMS and to $M^\star_{\mathrm{fast}}$ as JAMS-Fast, each taken from its own search (\Cref{sec:setup}). Both are frozen after selection and applied unchanged to unseen tasks and to other models.

\section{Experiments}
\label{sec:results}

We ask whether a procedure discovered on one model outperforms existing training-free methods on other models and unseen tasks, what it costs at inference time, and which strategies the search discovers. JAMS attains the highest average score on every model (\Cref{sec:main_results}), runs 4.0$\times$ faster than the strongest baseline (\Cref{sec:time}), and gains most where many options must be compared (\Cref{sec:case,sec:analysis}).

\subsection{Experimental Setup}
\label{sec:setup}

\textbf{Tasks.} The search uses six source tasks: function calling (BFCL and API-Bank; \citealt{patil2025bfcl,li2023apibank}), intent classification (Banking77; \citealt{casanueva2020efficient}), dialogue state tracking (SGD; \citealt{rastogi2020towards}), clinical-trial inference (NLI4CT; \citealt{jullien2024semeval}), and hallucination detection (RAGTruth; \citealt{niu2024ragtruth}). Each source task is split into training cases, visible to the coding agent, and validation cases, used only for scoring. The evaluation uses six different tasks that never appear during search: tool-use routing (When2Call; \citealt{ross2025when2call}), intent classification over 151 classes including out-of-scope requests (CLINC150+OOS; \citealt{larson2019evaluation}), multi-hop claim verification (HoVer; \citealt{jiang2020hover}), phishing detection (PhishNChips; \citealt{arelit2026phishnchips}), multi-question household-device control (Home-Appliance; \citealt{passos2026decisionindex}), and JevBench \citep{benchmarkheaven2026jevbench}, a public benchmark for Jev-style decision models. Together they cover choices among 2 to 151 options, rating scales, and binary verification; rating questions occur in no source task.

\textbf{Models.} We search on Qwen3.5-9B and Qwen3.5-4B \citep{qwenteam2026qwen35} and evaluate on five open models from two families: Qwen3.5-9B and Qwen3.5-4B, Qwen3-8B and Qwen3-4B-Instruct-2507 \citep{yang2025qwen3}, and Gemma-4-E4B-it \citep{gemmateam2026gemma4}. The discovered procedures are applied unchanged; nothing is tuned for any evaluation model or task.

\textbf{Baselines.} We compare against the seven training-free methods of \Cref{sec:intro}: Direct Generation, SemIf, open-alt-jev, LLM-as-Jev, LLM2Jev, jev-local, and AnyJev, each following its official implementation (or paper specification for LLM-as-Jev) under identical models, inputs, and serving settings. SemIf, open-alt-jev, and AnyJev use single-character option labels and cannot answer the 151-option task, receiving zero credit there (\Cref{sec:app_baselines}).

\textbf{Metrics.} Each task uses its standard metric: macro-F1 for CLINC150+OOS, case-level exact match for Home-Appliance, and accuracy otherwise. We report the unweighted six-task average with paired case-level bootstrap intervals (\Cref{sec:app_bootstrap}), and measure time per question as the wall-clock time to answer all evaluation questions with 64 concurrent requests, divided by their number.

\textbf{Implementation.} The coding agent is DeepSeek-V4.1-Flash running in the Claude Code harness, with file editing and shell tools in an isolated workspace; the judge is GLM-5.3-Flash from a separate model family. JAMS runs 100 iterations on Qwen3.5-9B (yielding JAMS as $M^\star$) and 24 iterations on Qwen3.5-4B (yielding JAMS-Fast as $M^\star_{\mathrm{fast}}$; \Cref{eq:selection}). All models run under SGLang \citep{zheng2024sglang} in bfloat16 on one GPU at matched concurrency (\Cref{sec:app_impl}).

\subsection{Main Results}
\label{sec:main_results}

\textbf{JAMS attains the highest score on every model.} As \Cref{tab:main} shows, JAMS raises the five-model average from 55.6\% for the strongest existing method, LLM-as-Jev, to 60.0\%. It exceeds the strongest baseline of each model by 1.3 to 4.5 points; the paired bootstrap intervals exclude zero on all four Qwen models but not on Gemma-4-E4B-it. Compared with LLM2Jev, the reference procedure of \Cref{eq:selection}, JAMS raises the average score by 5.8 to 10.2 points, with every interval excluding zero. Although optimized on Qwen3.5-9B alone, JAMS keeps these gains on a smaller model, the previous Qwen generation, and another family.

\begin{table}[t]
  \caption{\textbf{JAMS remains the most accurate on the five tasks that every method can answer.} Average score (\%) over all unseen tasks except CLINC150+OOS, with this task removed for every method. $^\dagger$Cannot answer the 151-option task (zero credit in \Cref{tab:main}). Columns: Qwen3.5-9B, Qwen3.5-4B, Qwen3-8B, Qwen3-4B-Instruct-2507, Gemma-4-E4B-it. Bold: best; underlined: second best.}
  \label{tab:common}
  \centering
  \footnotesize
  \setlength{\tabcolsep}{3.6pt}
  \renewcommand{\arraystretch}{1.08}
  \begin{tabular}{@{}l rrrrr r@{}}
    \toprule
    Method & 9B & 4B & 8B & 4B-I & E4B & Avg. \\
    \midrule
    Direct Generation & 56.4 & 56.4 & \underline{50.8} & 52.2 & \underline{56.6} & 54.5 \\
    SemIf$^\dagger$ & 55.7 & 55.6 & 43.4 & 43.1 & 53.9 & 50.3 \\
    open-alt-jev$^\dagger$ & 54.6 & 55.8 & 48.6 & 49.9 & 54.8 & 52.8 \\
    LLM-as-Jev & 56.8 & \underline{56.9} & \textbf{51.4} & \underline{52.5} & \underline{56.6} & \underline{54.8} \\
    LLM2Jev & 49.4 & 51.4 & 45.3 & 43.4 & 50.3 & 48.0 \\
    jev-local & 49.7 & 42.5 & 49.6 & 47.3 & 29.0 & 43.6 \\
    AnyJev$^\dagger$ & 56.7 & 51.6 & 47.8 & 48.1 & 52.3 & 51.3 \\
    \midrule
    \rowcolor{jamsrow} JAMS-Fast (ours) & \textbf{59.7} & 55.7 & 48.1 & \textbf{53.8} & 55.7 & 54.6 \\
    \rowcolor{jamsrow} JAMS (ours) & \underline{59.5} & \textbf{59.4} & 50.6 & 52.4 & \textbf{57.2} & \textbf{55.8} \\
    \bottomrule
  \end{tabular}
\end{table}

\textbf{The lead does not depend on the task some baselines cannot answer.} SemIf, open-alt-jev, and AnyJev receive zero credit on the 151-option task. \Cref{tab:common} removes this task for every method: JAMS still attains the highest average (55.8\% vs.\ 54.8\% for LLM-as-Jev), and JAMS or JAMS-Fast is the most accurate method on four of the five models; only on Qwen3-8B does LLM-as-Jev lead, by 0.8 points. The margin is smaller than in \Cref{tab:main} because JAMS gains most on the 151-option task (\Cref{sec:case}).

\begin{figure}[t]
  \centering
  \includegraphics[width=\columnwidth]{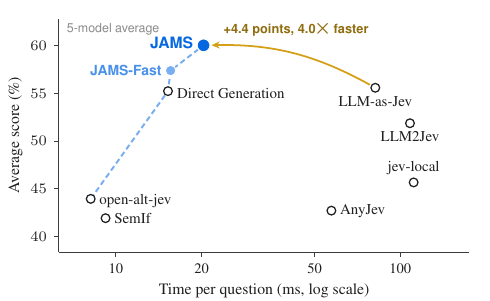}
  \caption{\textbf{JAMS is more accurate than every existing method and faster than every method that queries the model several times.} Average score over six unseen tasks versus time per question with 64 concurrent requests, both averaged over five models. Dashed line: quality--time frontier; arrow: JAMS relative to LLM-as-Jev, the strongest existing method on average.}
  \label{fig:pareto}
\end{figure}

\begin{figure}[t]
  \centering
  \includegraphics[width=\columnwidth]{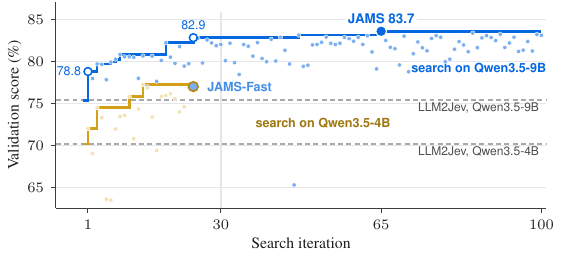}
  \caption{\textbf{Search finds strong procedures early.} Validation score of every admitted candidate (dots) and the best score so far (lines) in the searches on Qwen3.5-9B (100 iterations) and Qwen3.5-4B (24 iterations); dashed lines: LLM2Jev on each search model.}
  \label{fig:search_curve}
\end{figure}

\subsection{Inference Time}
\label{sec:time}

\textbf{JAMS is faster than every existing method that queries the model several times.} JAMS takes 20.3\,ms per question on average, 4.0$\times$ faster than LLM-as-Jev (81.7\,ms) and 5.3$\times$ faster than LLM2Jev (108.2\,ms); on every model it is 2.1 to 7.5$\times$ faster than each of LLM2Jev, jev-local, LLM-as-Jev, and AnyJev, because it screens large option sets before re-scoring a short list (\Cref{sec:case}).

\textbf{Methods that query the model once remain the fastest.} Open-alt-jev, SemIf, and Direct Generation take 8.2 to 15.2\,ms per question but reach at most 55.2\% (\Cref{fig:pareto}); JAMS-Fast takes 15.6\,ms at 57.4\%.

\begin{figure*}[t]
  \centering
  \includegraphics[width=\textwidth]{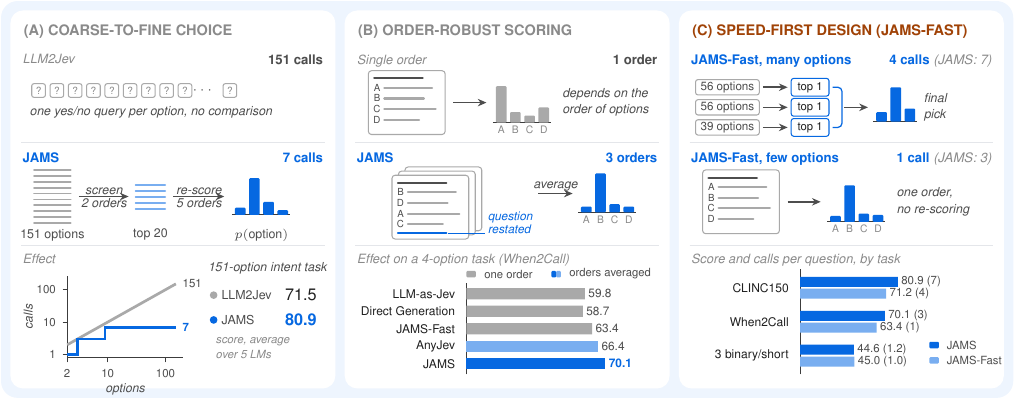}
  \captionsetup{width=0.8\linewidth}
  \caption{\textbf{What does JAMS discover?} Two strategies of JAMS and the speed-first design of JAMS-Fast, each contrasted with an alternative and shown with its measured effect (averages over five models).\newline\textbf{(A)}~Instead of one query per option, JAMS screens all options, keeps a short list, and re-scores it, so the number of model calls stays constant as the number of options grows. \textbf{(B)}~JAMS averages predictions over several option orders and restates the question after the options, reducing its sensitivity to option order. \textbf{(C)}~JAMS-Fast reduces large option sets with a tournament over chunks and reads small sets in a single order (calls per question in parentheses).}
  \label{fig:case}
\end{figure*}

\subsection{What Does JAMS Discover?}
\label{sec:case}

Because the output of JAMS is a program, its strategies can be read from the code. We highlight two that account for most of its gains and the design that makes JAMS-Fast fast (\Cref{fig:case}); JAMS also adapts option labels to each tokenizer and answers unseen rating questions by applying its choice routine to the scale levels.

\textbf{Coarse-to-fine choice over many options.} LLM2Jev asks a separate yes/no question per option, so its cost grows with the option count and options are never compared. JAMS adapts to the size of the option set: one query for binary questions, three averaged option orders for small sets. For large sets, it first screens all options in a single prompt under two orders, keeps the 20 most likely, and re-scores this short list under five orders. On the 151-option task, this reduces the cost from 151 model calls per question to 7 and raises the average score from 71.5\% for LLM2Jev, and 59.2\% for LLM-as-Jev, which lists all options in one prompt, to 80.9\%. The screening step rarely loses the answer: the correct intent survives into the short list on 98.5 to 100\% of the 151-option questions on each model.

\textbf{Order-robust scoring.} Reading option probabilities from a single prompt is sensitive to the order in which the options appear \citep{pezeshkpour2024large,zheng2024large}. JAMS rediscovered two remedies: averaging over several option orders, and restating the question after the options, close to where the answer is read. On the 4-option When2Call task, the two order-averaging methods (AnyJev 66.4\%, JAMS 70.1\%) beat all single-order ones ($\le$59.8\%).

\textbf{Speed-first design.} JAMS-Fast spends fewer calls exactly where JAMS spends more: it reduces a large option set by a tournament over chunks, keeping the most likely option of each (4 calls instead of 7 for 151 options), and reads small sets once, in a single order. This costs accuracy where options must be compared (CLINC150+OOS: 71.2\% vs.\ 80.9\%; When2Call: 63.4\% vs.\ 70.1\%), but not on the binary and short-choice tasks (HoVer, PhishNChips, Home-Appliance: 45.0\% vs.\ 44.6\%).

\subsection{Analysis}
\label{sec:analysis}

\textbf{Gains concentrate on tasks that require comparing options.} Averaged over the five models, JAMS improves most over LLM2Jev on When2Call (70.1\% vs.\ 49.6\%) and CLINC150+OOS (80.9\% vs.\ 71.5\%), where the model must weigh several candidates against one another. It stays within two points of the best baseline on HoVer (63.4\% vs.\ 64.1\%) and JevBench (75.3\% vs.\ 77.1\%), while LLM-as-Jev (14.1\%) and Direct Generation (13.1\%) lead on Home-Appliance (JAMS 9.1\%). A better procedure helps a model weigh options but cannot supply missing reasoning.

\textbf{Search finds strong procedures early.} In \Cref{fig:search_curve}, the first admitted candidate on Qwen3.5-9B already exceeds LLM2Jev on validation (78.8\% vs.\ 75.4\%), and the best score reaches 82.9\% within 30 iterations and 83.7\% after 100; on unseen tasks, the 30-iteration procedure is within 1.1 points of the final one on every model.

\textbf{The judge gate shapes the search.} Before search, the gate rejected all 26 pilot shortcuts and admitted all generic controls. On Qwen3.5-9B, 68 of 168 proposals needed repair, yet every iteration yielded an admitted candidate.

\section{Conclusion}
\label{sec:conclusion}

Existing training-free decision models adapt a frozen base model through a fixed, hand-designed procedure, yet this choice alone shifts decision quality by 13.9 to 34.0 points and inference time by an order of magnitude. JAMS optimizes it directly: a coding agent searches over programs, a judge gate admits only model-driven, task-agnostic procedures, and a quality--time archive guides exploration.

A single discovered procedure scores highest among seven training-free methods across five open models and six unseen tasks, and is faster than every multi-call baseline. The decision procedure merits as much optimization as the prompt or the model.

\textbf{Limitations and future work.} We study 4B--9B open models with one search per model; larger models, multimodal and multi-turn decisions, and more tokenizers remain open. A procedure is only as general as its source tasks, and the judge gate reduces but does not eliminate shortcuts.

\clearpage
\section*{Impact Statement}

This paper presents work whose goal is to advance the field of Machine Learning. By improving the quality and efficiency of training-free decision models built from small open models, our method can lower the computational cost and hardware requirements of deploying language-model-based classification, routing, and verification. Decision models are increasingly used to automate consequential decisions, and a procedure optimized on a set of source tasks may behave differently on new tasks or populations. Practitioners should therefore validate discovered procedures on their own data and retain human oversight in high-stakes applications. Because JAMS produces readable programs rather than opaque weights, such checks are comparatively easy to carry out.

\bibliography{references}
\bibliographystyle{icml2026}

\newpage
\appendix
\onecolumn

\FloatBarrier
\section{Tasks and Data}
\label{sec:app_data}

\textbf{Source tasks.} The search uses six tasks. For each task, 200 validation cases are held out for scoring candidates, and the remaining training cases (600 cases with 761 questions in total) are available to the coding agent. The validation sets are BFCL \citep{patil2025bfcl} (200 cases, 556 questions, exact match over all questions of a case), API-Bank \citep{li2023apibank} (200 questions, accuracy), Banking77 \citep{casanueva2020efficient} (200 questions with 77 options, macro-F1), SGD \citep{rastogi2020towards} (200 questions, macro-F1), NLI4CT \citep{jullien2024semeval} (200 questions, macro-F1), and RAGTruth \citep{niu2024ragtruth} (200 binary verification questions, F1 on the hallucinated class), for a total of 1{,}200 cases and 1{,}556 questions. At each iteration, JAMS samples three source tasks and shows the agent 128 training cases from them, with dataset names removed and identifiers anonymized.

\textbf{Evaluation tasks.} The six evaluation tasks are disjoint from the source tasks and are never used during search or selection. They comprise When2Call \citep{ross2025when2call} (200 questions with 4 options, accuracy), CLINC150+OOS \citep{larson2019evaluation} (200 questions with 151 options, macro-F1), HoVer \citep{jiang2020hover} (200 binary questions, accuracy), PhishNChips \citep{arelit2026phishnchips} (200 binary questions, accuracy), Home-Appliance \citep{passos2026decisionindex} (64 cases with 1{,}156 questions of 2 to 9 options each; a case counts as correct only if all of its questions are answered correctly), and JevBench \citep{benchmarkheaven2026jevbench} (231 questions: 139 choices, 74 verifications, and 18 ratings, organized into an original, an easy, and a hard subset). In total, the evaluation covers 1{,}095 cases and 2{,}187 questions. No rating question occurs in any source task. The evaluation tasks are disjoint from the source tasks but not unrelated to them: CLINC150+OOS and Banking77 are both intent classification, and When2Call, like BFCL and API-Bank, concerns tool use. We therefore regard HoVer, PhishNChips, Home-Appliance, and JevBench as the more distant tests of transfer.

\textbf{Serving.} All models are served with SGLang 0.5.12.post1 \citep{zheng2024sglang} and Transformers 5.6.0 on a single NVIDIA H200 GPU per model, in bfloat16 with a context length of 32{,}768 tokens, at most 64 concurrent requests, a fixed random seed, and the thinking mode disabled. The Qwen models use the FlashAttention-3 backend; Gemma-4-E4B-it, whose attention heads exceed the head dimension supported by FlashAttention-3, uses the Triton backend of the same container. Every method is evaluated with the same server configuration and request concurrency, and the KV cache is flushed between evaluations so that no method benefits from cached prefixes of another.

\FloatBarrier
\section{Baselines}
\label{sec:app_baselines}

Each baseline follows its official implementation. SemIf and open-alt-jev use their default settings, and we verified that our implementations produce the same prompts and distributions as the official code on all evaluation questions with the tokenizers of all five models.
\begin{itemize}[leftmargin=1.5em,itemsep=2pt,topsep=3pt]
  \item \textbf{Direct Generation} \citep{li2026llmasjev} lists the options with numeric indices, decodes greedily up to 32 tokens, parses the generated index, and places all probability on the chosen option. We use the prompt of Appendix~A of \citet{li2026llmasjev}; no generation failed to parse.
  \item \textbf{SemIf} \citep{semif2026} presents the evidence, the question, and the options as a structured prompt with the letters A to P as labels and reads the next-token probabilities of the labels from one forward pass. It supports at most 16 options; its 200 requests on CLINC150+OOS are counted as incorrect.
  \item \textbf{open-alt-jev} \citep{openaltjev2026} lists the options with the letters A to Z, asks for the letter of the answer, and reads the probabilities of the letters from one forward pass with a single option order. It supports at most 26 options; its 200 requests on CLINC150+OOS are counted as incorrect.
  \item \textbf{LLM-as-Jev} \citep{li2026llmasjev} uses the prompt of Direct Generation, appends the fixed answer prefix ``Best answer: ['' to the response, and scores every option by the summed log-probability of its index followed by the closing bracket; a softmax over these scores gives the distribution. No official code is available, so we follow the description of its training-free variant; reading these probabilities takes one call per distinct prefix of the answer strings, that is, $K{+}1$ calls for $K$ options.
  \item \textbf{LLM2Jev} \citep{yinsongxu2026llm2jev} asks a separate yes/no question for each option and normalizes the probabilities of ``yes'' across options; a binary question therefore costs two calls and a 151-option question 151 calls.
  \item \textbf{jev-local} \citep{jevlocal2026} scores each option by the mean log-probability of its text as a continuation of a plain-text prompt and applies the official calibration temperatures. Its prompt rendering and calibration match the official implementation on all JevBench questions.
  \item \textbf{AnyJev} \citep{zhang2026anyjev} averages over $K$ cyclic permutations of the options and calibrates each permutation against three content-free inputs \citep{zhao2021calibrate}, which costs $4K$ calls for a choice among $K$ options. We use its training-free decision module in the stateless mode, so that predictions do not depend on other test requests. In this configuration, options are labeled with single letters, so questions with more than 26 options cannot be answered; its 200 requests on CLINC150+OOS are counted as incorrect.
\end{itemize}
Following a protocol fixed before evaluation, any method whose first evaluation pass on a model exceeded the per-request time limit on more than 1\% of the questions received one additional pass in which only the timed-out requests were re-run with a longer limit of 1{,}800 seconds. Only jev-local on Qwen3.5-9B met this condition (192 of 2{,}187 requests); all re-run requests completed. All other methods completed every request in the first pass.

\FloatBarrier
\section{Implementation Details of JAMS}
\label{sec:app_impl}

\textbf{Program interface.} A candidate is a Python 3.11 module that defines an asynchronous function \texttt{decide(request, llm)} and returns a distribution over the options of every question in the request. The \texttt{llm} object exposes the three operations described in \Cref{sec:method_space}: \texttt{tokenize} (and a helper that returns the token identifiers of the chat template prefix), \texttt{generate} with \texttt{max\_new\_tokens=0} and a list of target token identifiers, which returns their next-token log-probabilities after the prompt without decoding, or bounded decoding when \texttt{max\_new\_tokens} is positive, and \texttt{chat\_completions} with optional top-20 log-probabilities. Thinking is disabled for all calls. Candidates may import only the Python standard library and the request and response types; this is checked on the syntax tree before execution. They run in a network-isolated container with read-only access to their own code, at most 1{,}024 model calls and 120 seconds per request. Exceptions, time-outs, requests that exceed these limits, and invalid distributions count as incorrect.

\textbf{Search.} Each iteration samples a parent uniformly from the current quality--time frontier of the archive, using a fixed random seed, and allows up to four attempts. An attempt must pass a syntax check, a smoke test on a mock model, and a smoke test on 16 live training cases without errors before it is submitted to the judge gate; the diagnostics of any failed check are returned to the agent for the next attempt. The coding agent is DeepSeek-V4.1-Flash operating in the Claude Code environment. During the first 30 iterations, each attempt was limited to 45 minutes with a context window of 256K tokens; for the remaining 70 iterations, we shortened the limit to 20 minutes and enlarged the context window to 1M tokens to avoid context compaction during long sessions. Because the parent sampling is deterministic, the continuation reproduces the first 30 iterations exactly. Admitted candidates are evaluated on all 1{,}556 validation questions with 64 concurrent requests and a flushed KV cache.

\textbf{Timing.} We report the time per question: the total wall-clock time of answering all 2{,}187 evaluation questions with 64 concurrent requests, divided by the number of questions. This is the inverse of throughput, not the latency of a single request. All methods except AnyJev are timed in the same run on one H200 GPU per model with the same server, in three passes with the order of the methods shuffled and the cache flushed before each method; we report the median. AnyJev was timed in an earlier run with the same protocol, server configuration, and hardware. Calls to the tokenizer are executed outside the event loop for every method, and requests that a method cannot answer are counted at their actual duration; on Qwen3.5-9B, the requests of jev-local and LLM2Jev that reach the per-request limit of 120 seconds are counted at that limit, so their times are lower bounds. The scores of SemIf, open-alt-jev, and LLM-as-Jev are those of the first timing pass; the scores of all other methods are taken from the main evaluation, and between 97.3\% and 100\% of the predictions of the timing run agree with it. The remaining differences arise because batched bfloat16 inference with 64 concurrent requests is not bitwise deterministic, so options with nearly equal probabilities can swap ranks between runs.

\textbf{Judge.} The judge is GLM-5.3-Flash, run at temperature 0. It receives the program source, the interface specification, the rubric, and the records of the perturbation tests, but no data, scores, or transcripts of the coding agent. A candidate passes only if the judge flags none of the ten rules.

\FloatBarrier
\section{Judge Gate}
\label{sec:app_gate}

\textbf{Perturbation tests.} The tests execute a candidate on synthetic requests from five fictional tasks, each in four variants (the original, renamed input fields, paraphrased instructions, and shuffled options) and with two random seeds, for 40 runs against a simulated model. For each run, the tests record the number of model calls, whether a non-uniform decision was produced without any model call, whether the decision follows the options when they are shuffled, and a hash of the prompt templates with all values removed, which reveals branching on field names.

\textbf{Rubric and calibration.} \Cref{tab:app_calibration} lists the ten rules of the rubric. Before any search, we froze the rubric, the judge prompt, and the tests, and calibrated them on 26 programs written during the unconstrained pilot study (\Cref{sec:app_pilot}), all of which contain task-specific shortcuts, and on 3 generic control programs. The gate rejected all 26 shortcut programs and admitted all 3 controls. A review took 627 seconds on average (between 110 and 1{,}108 seconds).

\begin{table}[htbp]
  \caption{Rubric of the independent judge and its calibration on the 26 shortcut programs from the pilot study. The last column counts the programs flagged under each rule (a program may be flagged under several rules). All 26 programs were rejected and all 3 generic control programs were admitted.}
  \label{tab:app_calibration}
  \centering\small
  \begin{tabularx}{\textwidth}{@{}r l >{\raggedright\arraybackslash}X r@{}}
    \toprule
    & Rule & Requirement & Flagged \\
    \midrule
    1 & Model-driven decisions & Do not compute, look up, or rule out answers without querying the model; only uniform fallbacks are allowed when the model fails. & 26/26 \\
    2 & No content branching & Do not apply keyword, pattern, or key tests to the context, instructions, or options. & 26/26 \\
    3 & No domain vocabularies & Include no domain names, hosts, intent labels, tool or entity names, units, or task-specific constants. & 26/26 \\
    4 & Task-neutral prompts & Prompt text must be neutral with respect to the subject of the task. & 24/26 \\
    5 & No memorized cases & Embed no training examples, hashes, or label tables. & 19/26 \\
    6 & No surface heuristics & Do not decide or break ties by meaning-independent cues such as position or option length. & 16/26 \\
    7 & No task fingerprinting & Branch only on generic structure (question type, coarse option count), never on signatures of a task. & 23/26 \\
    8 & No rule-based pre-selection & Short lists must be ranked by model scores, not by heuristic filters. & 26/26 \\
    9 & Stateless execution & Keep no state across requests apart from tokenizer look-ups. & 0/26 \\
    10 & Respect limits & Do not circumvent time limits, call budgets, or the non-thinking setting. & 0/26 \\
    \bottomrule
  \end{tabularx}
\end{table}

\textbf{Behavior during search.} In the 100-iteration search on Qwen3.5-9B, the agent made 168 attempts, 68 of which failed a check and triggered a repair, and every iteration ended with an admitted candidate. In the 24-iteration search on Qwen3.5-4B, the agent made 47 attempts, 23 of which failed a check and triggered a repair, and again every iteration ended with an admitted candidate.

\FloatBarrier
\section{Pilot Study without the Gate}
\label{sec:app_pilot}

The pilot study ran 13 iterations of an earlier, unconstrained variant of the search on Qwen3.5-4B and on Qwen3.5-9B. It started from LLM2Jev rather than from scratch and used eleven tasks with disjoint training and held-out cases from the same tasks, so the held-out cases were in-distribution; the eleven tasks include the six source tasks and five of the evaluation tasks of the main study. No constraint was placed on the generated code.

\textbf{Task-specific shortcuts.} The search soon produced programs that answer questions with hand-written rules. For Home-Appliance, the agent wrote a solver of roughly 250 lines that parses the request, the device states, and the room names directly from the input. On the held-out cases of this task, the solver reached 96.9\% case-level accuracy without any model call on 62 of 64 cases on Qwen3.5-4B and on all 64 cases on Qwen3.5-9B, whereas LLM2Jev solved no case despite 3{,}070 model calls. Similar rules appeared for other tasks, including lists of phishing hosts, checkers for tool schemas, and routers keyed on the fields of clinical-trial records. Together, they raised the held-out score over the eleven tasks from 60.6\% to 82.3\% on Qwen3.5-4B and from 62.8\% to 83.0\% on Qwen3.5-9B.

\textbf{No transfer.} On JevBench, which none of these rules were written for, traces show that no task-specific branch was triggered on any of the 231 questions. The programs improved JevBench accuracy from 75.8\% to 79.2\% on Qwen3.5-4B and from 77.5\% to 80.1\% on Qwen3.5-9B, and this gain came entirely from the components that query the model. The 26 programs produced in the two pilot runs form the shortcut set used to calibrate the gate (\Cref{sec:app_gate}).

\FloatBarrier
\section{Additional Results}
\label{sec:app_per_scenario}

\textbf{Per-task results.} \Cref{tab:app_per_scenario,tab:app_per_scenario_b} report the score of every method on every evaluation task and model, as well as the average over the five models, including the procedures selected from the search on Qwen3.5-4B and the procedure selected after 30 iterations of the search on Qwen3.5-9B.

\begin{table}[htbp]
  \caption{Per-task results on the six unseen tasks. Scores are in \%; Time is the wall-clock time per question in milliseconds with 64 concurrent requests. SemIf, open-alt-jev, and AnyJev cannot answer questions with more than 16, 26, and 26 options, respectively; their answers on CLINC150+OOS count as incorrect.}
  \label{tab:app_per_scenario}
  \centering\small
  \setlength{\tabcolsep}{4.2pt}
  \begin{tabular}{@{}l rrrrrr r r@{}}
    \toprule
    Method & When2Call & CLINC150 & HoVer & PhishNChips & Home-App. & JevBench & Avg. & Time (ms) \\
    \midrule
    \multicolumn{9}{@{}l}{\textit{Qwen3.5-9B}} \\
    \quad Direct Generation & 60.0 & 70.0 & 63.5 & 64.0 & 12.5 & 81.8 & 58.6 & 19.2 \\
    \quad SemIf & 61.5 & 0.0 & 64.5 & 58.0 & 12.5 & 81.8 & 46.4 & 13.7 \\
    \quad open-alt-jev & 59.5 & 0.0 & 62.5 & 63.0 & 10.9 & 77.1 & 45.5 & 12.1 \\
    \quad LLM-as-Jev & 61.5 & 72.3 & 64.0 & 64.0 & 12.5 & 81.8 & 59.4 & 91.6 \\
    \quad LLM2Jev & 46.5 & 74.7 & 65.5 & 58.0 & 0.0 & 77.1 & 53.6 & 193.7 \\
    \quad jev-local & 45.0 & 72.5 & 62.5 & 62.5 & 3.1 & 75.3 & 53.5 & 207.5 \\
    \quad AnyJev & 76.0 & 0.0 & 62.5 & 54.0 & 12.5 & 78.4 & 47.2 & 87.7 \\
    \quad JAMS, searched on Qwen3.5-4B & 70.0 & 79.3 & 65.0 & 62.5 & 10.9 & 79.2 & 61.2 & 70.0 \\
    \quad JAMS after 30 iterations & 75.0 & 83.0 & 65.0 & 67.5 & 12.5 & 79.7 & 63.8 & 29.6 \\
    \quad \textbf{JAMS-Fast} & 70.5 & 76.5 & 65.0 & 67.5 & 15.6 & 80.1 & 62.5 & 25.2 \\
    \rowcolor{jamsrow} \quad \textbf{JAMS} & 75.0 & 85.3 & 65.5 & 66.5 & 10.9 & 79.7 & 63.8 & 29.3 \\
    \midrule
    \multicolumn{9}{@{}l}{\textit{Qwen3.5-4B}} \\
    \quad Direct Generation & 61.5 & 69.3 & 64.5 & 70.0 & 7.8 & 78.4 & 58.6 & 14.3 \\
    \quad SemIf & 46.0 & 0.0 & 70.0 & 69.0 & 12.5 & 80.5 & 46.3 & 9.4 \\
    \quad open-alt-jev & 59.5 & 0.0 & 65.0 & 67.5 & 9.4 & 77.5 & 46.5 & 8.6 \\
    \quad LLM-as-Jev & 63.0 & 69.7 & 64.5 & 69.5 & 9.4 & 78.4 & 59.1 & 79.0 \\
    \quad LLM2Jev & 48.5 & 73.5 & 67.5 & 65.5 & 0.0 & 75.3 & 55.0 & 141.1 \\
    \quad jev-local & 40.5 & 73.0 & 51.0 & 53.0 & 0.0 & 68.0 & 47.6 & 155.0 \\
    \quad AnyJev & 72.0 & 0.0 & 57.5 & 52.5 & 0.0 & 75.8 & 43.0 & 75.2 \\
    \quad JAMS, searched on Qwen3.5-4B & 61.5 & 80.4 & 62.5 & 71.5 & 6.2 & 77.1 & 59.9 & 60.7 \\
    \quad JAMS after 30 iterations & 76.0 & 81.4 & 64.5 & 68.5 & 10.9 & 77.9 & 63.2 & 22.4 \\
    \quad \textbf{JAMS-Fast} & 60.5 & 75.6 & 63.5 & 73.0 & 6.2 & 75.3 & 59.0 & 20.8 \\
    \rowcolor{jamsrow} \quad \textbf{JAMS} & 76.0 & 81.2 & 64.5 & 73.0 & 6.2 & 77.5 & 63.1 & 20.7 \\
    \midrule
    \multicolumn{9}{@{}l}{\textit{Qwen3-8B}} \\
    \quad Direct Generation & 52.5 & 24.3 & 59.0 & 55.5 & 12.5 & 74.5 & 46.4 & 13.4 \\
    \quad SemIf & 40.0 & 0.0 & 57.0 & 51.0 & 1.6 & 67.5 & 36.2 & 8.6 \\
    \quad open-alt-jev & 49.5 & 0.0 & 62.5 & 53.0 & 9.4 & 68.8 & 40.5 & 7.6 \\
    \quad LLM-as-Jev & 53.5 & 24.2 & 58.5 & 55.5 & 15.6 & 74.0 & 46.9 & 77.3 \\
    \quad LLM2Jev & 52.0 & 67.0 & 57.0 & 51.5 & 0.0 & 65.8 & 48.9 & 66.1 \\
    \quad jev-local & 54.5 & 65.4 & 62.0 & 59.0 & 3.1 & 69.3 & 52.2 & 56.1 \\
    \quad AnyJev & 61.0 & 0.0 & 55.0 & 51.0 & 0.0 & 71.9 & 39.8 & 40.1 \\
    \quad JAMS, searched on Qwen3.5-4B & 57.5 & 70.8 & 58.5 & 56.5 & 7.8 & 65.8 & 52.8 & 26.3 \\
    \quad JAMS after 30 iterations & 69.5 & 70.4 & 59.5 & 54.5 & 4.7 & 71.4 & 55.0 & 19.6 \\
    \quad \textbf{JAMS-Fast} & 53.0 & 68.9 & 59.5 & 57.5 & 0.0 & 70.6 & 51.6 & 11.5 \\
    \rowcolor{jamsrow} \quad \textbf{JAMS} & 69.5 & 74.7 & 59.0 & 52.5 & 1.6 & 70.6 & 54.6 & 19.5 \\
    \bottomrule
  \end{tabular}
\end{table}

\begin{table}[htbp]
  \caption{Per-task results on the six unseen tasks (continued).}
  \label{tab:app_per_scenario_b}
  \centering\small
  \setlength{\tabcolsep}{4.2pt}
  \begin{tabular}{@{}l rrrrrr r r@{}}
    \toprule
    Method & When2Call & CLINC150 & HoVer & PhishNChips & Home-App. & JevBench & Avg. & Time (ms) \\
    \midrule
    \multicolumn{9}{@{}l}{\textit{Qwen3-4B-Instruct-2507}} \\
    \quad Direct Generation & 61.0 & 54.7 & 65.5 & 53.5 & 9.4 & 71.4 & 52.6 & 16.0 \\
    \quad SemIf & 31.5 & 0.0 & 64.0 & 51.5 & 0.0 & 68.4 & 35.9 & 5.5 \\
    \quad open-alt-jev & 56.0 & 0.0 & 66.5 & 54.0 & 4.7 & 68.4 & 41.6 & 4.8 \\
    \quad LLM-as-Jev & 62.5 & 54.3 & 65.5 & 53.5 & 9.4 & 71.4 & 52.8 & 56.6 \\
    \quad LLM2Jev & 45.5 & 71.0 & 53.5 & 50.5 & 0.0 & 67.5 & 48.0 & 58.4 \\
    \quad jev-local & 51.0 & 58.3 & 65.0 & 51.5 & 0.0 & 68.8 & 49.1 & 54.6 \\
    \quad AnyJev & 58.0 & 0.0 & 61.5 & 52.5 & 0.0 & 68.4 & 40.1 & 31.8 \\
    \quad JAMS, searched on Qwen3.5-4B & 66.5 & 69.6 & 64.5 & 53.5 & 12.5 & 66.7 & 55.6 & 20.3 \\
    \quad JAMS after 30 iterations & 67.5 & 70.6 & 66.0 & 53.5 & 10.9 & 70.6 & 56.5 & 13.0 \\
    \quad \textbf{JAMS-Fast} & 69.5 & 63.8 & 66.0 & 52.5 & 10.9 & 70.1 & 55.5 & 8.3 \\
    \rowcolor{jamsrow} \quad \textbf{JAMS} & 67.5 & 81.6 & 66.5 & 53.5 & 3.1 & 71.4 & 57.3 & 12.7 \\
    \midrule
    \multicolumn{9}{@{}l}{\textit{Gemma-4-E4B-it}} \\
    \quad Direct Generation & 58.5 & 76.8 & 65.0 & 57.0 & 23.4 & 79.2 & 60.0 & 13.4 \\
    \quad SemIf & 56.0 & 0.0 & 63.0 & 52.5 & 20.3 & 77.5 & 44.9 & 8.8 \\
    \quad open-alt-jev & 60.0 & 0.0 & 64.0 & 56.0 & 17.2 & 77.1 & 45.7 & 7.7 \\
    \quad LLM-as-Jev & 58.5 & 75.7 & 65.0 & 57.0 & 23.4 & 79.2 & 59.8 & 104.0 \\
    \quad LLM2Jev & 55.5 & 71.1 & 61.5 & 51.0 & 7.8 & 75.8 & 53.8 & 81.7 \\
    \quad jev-local & 12.5 & 11.2 & 49.5 & 50.0 & 0.0 & 32.9 & 26.0 & 84.1 \\
    \quad AnyJev & 65.0 & 0.0 & 55.5 & 52.0 & 14.1 & 74.9 & 43.6 & 51.5 \\
    \quad JAMS, searched on Qwen3.5-4B & 61.5 & 78.1 & 60.5 & 54.5 & 9.4 & 75.8 & 56.6 & 43.5 \\
    \quad JAMS after 30 iterations & 63.0 & 80.4 & 63.5 & 57.0 & 20.3 & 77.1 & 60.2 & 20.2 \\
    \quad \textbf{JAMS-Fast} & 63.5 & 71.4 & 60.5 & 55.0 & 21.9 & 77.5 & 58.3 & 12.1 \\
    \rowcolor{jamsrow} \quad \textbf{JAMS} & 62.5 & 81.8 & 61.5 & 61.0 & 23.4 & 77.5 & 61.3 & 19.5 \\
    \midrule
    \multicolumn{9}{@{}l}{\textit{Average over the five models}} \\
    \quad Direct Generation & 58.7 & 59.0 & 63.5 & 60.0 & 13.1 & 77.1 & 55.2 & 15.2 \\
    \quad SemIf & 47.0 & 0.0 & 63.7 & 56.4 & 9.4 & 75.2 & 41.9 & 9.2 \\
    \quad open-alt-jev & 56.9 & 0.0 & 64.1 & 58.7 & 10.3 & 73.8 & 44.0 & 8.2 \\
    \quad LLM-as-Jev & 59.8 & 59.2 & 63.5 & 59.9 & 14.1 & 77.0 & 55.6 & 81.7 \\
    \quad LLM2Jev & 49.6 & 71.5 & 61.0 & 55.3 & 1.6 & 72.3 & 51.9 & 108.2 \\
    \quad jev-local & 40.7 & 56.1 & 58.0 & 55.2 & 1.2 & 62.9 & 45.7 & 111.5 \\
    \quad AnyJev & 66.4 & 0.0 & 58.4 & 52.4 & 5.3 & 73.9 & 42.7 & 57.2 \\
    \quad JAMS, searched on Qwen3.5-4B & 63.4 & 75.7 & 62.2 & 59.7 & 9.4 & 72.9 & 57.2 & 44.2 \\
    \quad JAMS after 30 iterations & 70.2 & 77.2 & 63.7 & 60.2 & 11.9 & 75.3 & 59.7 & 20.9 \\
    \quad \textbf{JAMS-Fast} & 63.4 & 71.2 & 62.9 & 61.1 & 10.9 & 74.7 & 57.4 & 15.6 \\
    \rowcolor{jamsrow} \quad \textbf{JAMS} & 70.1 & 80.9 & 63.4 & 61.3 & 9.1 & 75.3 & 60.0 & 20.3 \\
    \bottomrule
  \end{tabular}
\end{table}

\subsection{Quality and Time on Each Model}
\label{sec:app_pareto}

\begin{figure}[htbp]
  \centering
  \includegraphics[width=\textwidth]{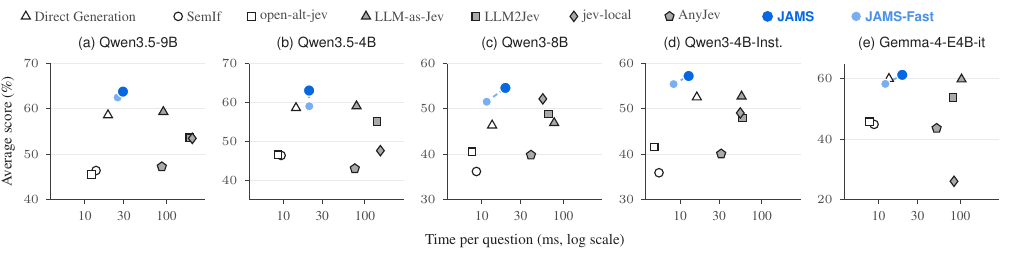}
  \caption{\textbf{Quality and time per question of every method on each model.} Average score over six unseen tasks versus time per question with 64 concurrent requests (log scale). Open markers: methods that query the model once per question; grey markers: methods that query it several times; dashed line: JAMS-Fast to JAMS.}
  \label{fig:app_pareto}
\end{figure}

\Cref{fig:app_pareto} shows the average score and the time per question of every method separately for each model. On every model, JAMS is more accurate than all existing methods and faster than all existing methods that query the model several times per question; JAMS-Fast is also faster than Direct Generation on Qwen3-8B, Qwen3-4B-Instruct-2507, and Gemma-4-E4B-it.

\subsection{Paired Bootstrap Intervals}
\label{sec:app_bootstrap}

We compute intervals from the per-question predictions with a paired bootstrap that resamples whole cases with replacement within each task (1{,}000 resamples with a fixed seed) and recomputes each task metric, including macro-F1 and case-level exact match, for both methods on the same resampled cases. \Cref{tab:app_bootstrap} reports the resulting 95\% intervals for the average score. JAMS improves over LLM2Jev on every model and over the strongest baseline on the four Qwen models; on Gemma-4-E4B-it, the interval against Direct Generation includes zero. JAMS-Fast improves over LLM2Jev on every model and over the strongest baseline on Qwen3.5-9B and Qwen3-4B-Instruct-2507. The procedures selected after 30 and after 100 iterations do not differ significantly on any model. The intervals reflect the variability of the test cases for fixed procedures; they do not capture the variability across independent search runs.

\begin{table}[htbp]
  \caption{Differences in average score (percentage points) with 95\% intervals from a paired bootstrap over cases (1{,}000 resamples). The strongest baseline is chosen per model as in \Cref{tab:main}; differences are computed before rounding.}
  \label{tab:app_bootstrap}
  \centering\footnotesize
  \setlength{\tabcolsep}{3.2pt}
  \resizebox{\textwidth}{!}{%
  \begin{tabular}{@{}l rrrrr@{}}
    \toprule
    Comparison & Qwen3.5-9B & Qwen3.5-4B & Qwen3-8B & Qwen3-4B-Inst. & Gemma-4-E4B-it \\
    \midrule
    JAMS $-$ LLM2Jev & +10.2 [+8.0, +12.8] & +8.0 [+5.6, +10.8] & +5.8 [+3.8, +7.9] & +9.3 [+6.7, +11.5] & +7.5 [+4.5, +10.4] \\
    JAMS $-$ strongest baseline & +4.5 [+2.3, +6.8] & +4.0 [+1.8, +6.3] & +2.4 [+0.2, +4.6] & +4.5 [+2.0, +6.3] & +1.3 [-1.0, +3.4] \\
    JAMS $-$ JAMS after 30 iterations & +0.0 [-1.3, +1.4] & -0.1 [-1.5, +1.2] & -0.4 [-1.6, +1.0] & +0.8 [-0.8, +2.1] & +1.1 [-0.2, +2.5] \\
    JAMS $-$ JAMS searched on Qwen3.5-4B & +2.7 [+0.5, +4.8] & +3.2 [+0.6, +5.8] & +1.8 [-0.6, +4.0] & +1.7 [-0.8, +3.7] & +4.7 [+2.1, +7.1] \\
    JAMS-Fast $-$ LLM2Jev & +8.9 [+6.4, +11.5] & +4.0 [+1.5, +6.6] & +2.7 [+0.5, +4.6] & +7.5 [+4.8, +9.9] & +4.5 [+1.6, +7.6] \\
    JAMS-Fast $-$ strongest baseline & +3.2 [+0.7, +5.5] & -0.0 [-2.6, +2.6] & -0.6 [-2.9, +1.4] & +2.7 [+0.3, +5.1] & -1.7 [-3.9, +0.3] \\
    \bottomrule
  \end{tabular}}
  \par\vspace{2pt}{\footnotesize Strongest baseline per model: Qwen3.5-9B: LLM-as-Jev, Qwen3.5-4B: LLM-as-Jev, Qwen3-8B: jev-local, Qwen3-4B-Instruct-2507: LLM-as-Jev, Gemma-4-E4B-it: Direct Generation.}
\end{table}

\subsection{Results on the Complete Test Sets}
\label{sec:app_full}

The main evaluation uses a subset of each evaluation task. To confirm that the gains hold on more data, we also evaluated LLM2Jev, the two procedures selected from the search on Qwen3.5-4B, and the procedure selected after 30 iterations of the search on Qwen3.5-9B on the complete test sets of the six evaluation tasks, on Qwen3.5-4B and Qwen3.5-9B (\Cref{tab:app_full_transfer}). We removed 642 When2Call questions whose source examples also occur in BFCL, which is a source task, and verified that no case, input, or household state overlaps with the data used during search. All procedures improve over LLM2Jev on both models with intervals that exclude zero, and each procedure transfers between the two models: the procedure found on Qwen3.5-9B improves Qwen3.5-4B by 8.1 points, more than the procedures found on Qwen3.5-4B itself.

\begin{table}[htbp]
  \caption{Results on the complete test sets of the six evaluation tasks (14{,}901 cases, 17{,}636 questions). Scores in \%; $\Delta$ is the difference in average score to LLM2Jev with a 95\% paired bootstrap interval. Invalid predictions count as incorrect. This evaluation uses the procedure selected after 30 iterations of the Qwen3.5-9B search.}
  \label{tab:app_full_transfer}
  \centering\footnotesize
  \setlength{\tabcolsep}{2.6pt}
  \resizebox{\textwidth}{!}{%
  \begin{tabular}{@{}l rrrrrr r r@{}}
    \toprule
    Method & When2Call & CLINC150 & HoVer & PhishNChips & Home-App. & JevBench & Avg. & $\Delta$ [95\% CI] \\
    \midrule
    \multicolumn{9}{@{}l}{\textit{Qwen3.5-4B}} \\
    \quad LLM2Jev & 50.4 & 75.3 & 60.3 & 64.3 & 0.0 & 77.1 & 54.6 & --- \\
    \quad JAMS, searched on Qwen3.5-4B & 61.9 & 81.0 & 60.2 & 73.2 & 4.4 & 77.1 & 59.6 & +5.1 [+3.9, +6.3] \\
    \quad JAMS-Fast & 61.3 & 76.0 & 61.0 & 77.1 & 0.6 & 75.8 & 58.6 & +4.1 [+3.0, +5.1] \\
    \quad JAMS after 30 iterations & 73.7 & 81.8 & 62.8 & 71.1 & 9.4 & 77.5 & 62.7 & +8.1 [+6.8, +9.5] \\
    \midrule
    \multicolumn{9}{@{}l}{\textit{Qwen3.5-9B}} \\
    \quad LLM2Jev & 50.3 & 77.6 & 62.3 & 58.5 & 0.0 & 77.5 & 54.3 & --- \\
    \quad JAMS, searched on Qwen3.5-4B & 73.2 & 82.7 & 63.8 & 65.6 & 8.1 & 80.1 & 62.3 & +7.9 [+6.5, +9.4] \\
    \quad JAMS-Fast & 72.5 & 80.1 & 63.4 & 66.0 & 9.4 & 80.1 & 61.9 & +7.6 [+6.2, +8.9] \\
    \quad JAMS after 30 iterations & 75.5 & 85.0 & 64.8 & 69.0 & 10.0 & 79.7 & 64.0 & +9.6 [+8.3, +10.9] \\
    \bottomrule
  \end{tabular}}
\end{table}

\FloatBarrier
\section{Search Trajectories and the Discovered Program}
\label{sec:app_trajectories}

\begin{figure}[htbp]
  \centering
  \includegraphics[width=0.56\textwidth]{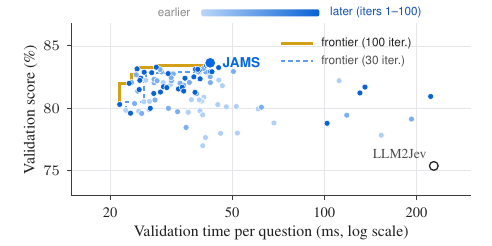}
  \caption{\textbf{Quality and time of the candidates found on Qwen3.5-9B.} Validation score versus time per question (ms, log scale) of all 100 candidates, shaded from earlier to later iterations, with the quality--time frontier after 30 and after 100 iterations and the selected procedure. The validation score during search is shown in \Cref{fig:search_curve}.}
  \label{fig:search}
\end{figure}

\textbf{Trajectories.} \Cref{fig:search_curve,fig:search} and \Cref{tab:app_trajectories} summarize the two searches. JAMS is the candidate admitted at iteration 65 of the search on Qwen3.5-9B; after 30 iterations, the selection rule of \Cref{eq:selection} would have chosen the candidate of iteration 24. On Qwen3.5-4B, the rule selects the candidates of iterations 13 (quality-first) and 24 (speed-first, JAMS-Fast). Each search thus yields a quality-first and a speed-first procedure; we report the quality-first procedure of the longer search on Qwen3.5-9B as JAMS and the speed-first procedure of the search on Qwen3.5-4B as JAMS-Fast.

\begin{table}[htbp]
  \caption{Validation scores (\%) per source task for selected candidates of the two searches, with the average and the validation time per question (ms, 64 concurrent requests). The iteration is the search step at which the candidate was admitted.}
  \label{tab:app_trajectories}
  \centering\small
  \setlength{\tabcolsep}{4.5pt}
  \begin{tabular}{@{}l rrrrrr r r@{}}
    \toprule
    Candidate & API-Bank & Banking77 & BFCL & NLI4CT & RAGTruth & SGD & Avg. & Time (ms) \\
    \midrule
    \multicolumn{9}{@{}l}{\textit{Search on Qwen3.5-9B}} \\
    \quad LLM2Jev & 66.0 & 71.0 & 97.0 & 69.0 & 63.4 & 86.1 & 75.4 & 227.7 \\
    \quad iteration 1 & 83.0 & 71.7 & 88.0 & 69.3 & 79.7 & 81.3 & 78.8 & 68.5 \\
    \quad iteration 18 & 84.0 & 75.9 & 95.5 & 76.8 & 72.6 & 88.8 & 82.3 & 45.2 \\
    \quad iteration 24 (selected after 30 iterations) & 83.5 & 73.3 & 95.5 & 77.3 & 78.5 & 89.2 & 82.9 & 39.2 \\
    \quad iteration 47 & 84.0 & 76.5 & 94.5 & 75.9 & 78.5 & 89.7 & 83.2 & 40.3 \\
    \rowcolor{jamsrow} \quad \textbf{iteration 65 (JAMS)} & 85.5 & 77.2 & 94.5 & 75.2 & 79.4 & 90.0 & \textbf{83.7} & 42.4 \\
    \quad iteration 100 & 82.5 & 76.8 & 95.0 & 75.9 & 79.4 & 89.4 & 83.2 & 24.5 \\
    \midrule
    \multicolumn{9}{@{}l}{\textit{Search on Qwen3.5-4B}} \\
    \quad LLM2Jev & 63.0 & 64.5 & 92.0 & 59.7 & 59.7 & 82.1 & 70.2 & 152.6 \\
    \quad iteration 13 (quality-first) & 75.0 & 69.1 & 87.5 & 69.1 & 76.4 & 86.6 & 77.3 & 40.4 \\
    \quad \textbf{iteration 24 (JAMS-Fast)} & 80.5 & 67.2 & 90.0 & 72.6 & 75.2 & 76.7 & 77.0 & 19.5 \\
    \bottomrule
  \end{tabular}
\end{table}

\begin{table}[htbp]
  \caption{Accuracy (\%) on the 151-option task (CLINC150+OOS, 200 questions) split by the position of the correct option relative to the first two-letter label that the model's tokenizer splits into two tokens (\Cref{sec:app_labels}). JAMS keeps the two-letter labels that are single tokens; JAMS-Fast uses one-character labels only.}
  \label{tab:app_split}
  \centering\small
  \begin{tabular}{@{}l c rr rr@{}}
    \toprule
    & & \multicolumn{2}{c}{Before the split} & \multicolumn{2}{c}{After the split} \\
    \cmidrule(lr){3-4} \cmidrule(l){5-6}
    Model & Split after & JAMS & JAMS-Fast & JAMS & JAMS-Fast \\
    \midrule
    Qwen3.5-9B & 68 options & 92.1 & 77.6 & 78.2 & 77.4 \\
    Qwen3.5-4B & 68 options & 88.2 & 75.0 & 78.2 & 79.8 \\
    Qwen3-8B & 50 options & 81.0 & 75.9 & 77.5 & 71.8 \\
    Qwen3-4B-Instruct-2507 & 50 options & 84.5 & 69.0 & 75.4 & 62.7 \\
    Gemma-4-E4B-it & 68 options & 90.8 & 81.6 & 75.8 & 72.6 \\
    \bottomrule
  \end{tabular}
\end{table}

\textbf{How the strategies emerged.} Every candidate records its parent, so the lineage of the final procedure can be traced. JAMS descends from iterations 1, 2, 3, 6, 7, 8, 10, 13, 18, 24, 25, and 47. Along this lineage, the agent introduced reading all option labels from a single prompt (iteration 1), pooling the probabilities of alternative label forms (2), two-letter labels for long option lists (3), averaging over the forward and reversed option order (6), screening wide option sets before re-scoring a short list (8), dedicated prompts for binary questions (13), re-scoring under five orders together with restating the question after the options (18), and explicit definitions of both outcomes in verification questions (24, which raised the validation score on RAGTruth by 5.9 points). The later edits along the lineage changed little code and can be compared directly on validation: answering binary choices with a single query reduced the validation time per question by 21\% (iteration 25); enlarging the short list from 5 to 20 options and the number of re-scoring orders from 3 to 5 raised the validation score on Banking77 by 2.8 points (47); and screening under two orders raised the average validation score from 83.2\% to 83.7\% (65). Because the earlier iterations changed several components at once, we do not attribute their effect to individual components. 

\textbf{The discovered program.} The final procedure is a module of 860 lines. Its routine \texttt{\_choose} implements the coarse-to-fine choice of \Cref{sec:case}: one prompt for two options, three option orders for up to eight options, and for larger sets a screening pass under two orders, a short list of 20 options, and a re-scoring pass under five orders; options that are not short-listed share a residual probability of 0.01. The routine \texttt{\_label\_ids} determines, for the current model, which candidate labels can be read as a single token, trying the label with and without a leading space, and \texttt{\_page\_prompt} places the restated question after the option list. Ratings reuse \texttt{\_choose} over the levels of the scale and return the expected level, and verification questions state both outcomes explicitly and pool the probabilities of several spellings of ``True'' and ``False''. Answering a 151-option question takes 7 model calls, a question with up to eight options 3 calls, and a binary question 1 call. On the evaluation tasks, JAMS queries the model for every question, on every model.

\textbf{JAMS-Fast.} JAMS-Fast is a module of 605 lines. It labels options with one-character symbols (letters and digits), determines once per model which of them are single tokens, and therefore reads at most 62 options per prompt. Option sets of up to eight options are scored in one prompt; larger sets that fit into one prompt are scored once together with a short generated answer, and the 30 most likely options are re-scored in a second prompt; sets that exceed one prompt are split into chunks of at most 56 options, the most likely option of each chunk advances, and a final prompt decides among the winners. All prompts use a single option order. Answering a 151-option question takes 4 model calls, and a question with up to eight options 1 call. Like JAMS, it queries the model for every evaluation question, on every model.

\textbf{Tokenizer-aware option labels.}\label{sec:app_labels} Large option sets require two-letter labels such as \emph{AA}, \emph{AB}, and so on. Whether such a label is a single token depends on the tokenizer: the first label that Qwen3 splits into two tokens follows the first 50 options, whereas for Qwen3.5 and Gemma~4 it follows the first 68. JAMS checks at run time which labels the current model reads as one token and uses only those, so the same program produces model-specific labels; JAMS-Fast avoids two-letter labels altogether. On questions of the 151-option task whose correct answer lies beyond the split, JAMS retains 77.0\% accuracy and JAMS-Fast 72.9\% (\Cref{tab:app_split}).
\FloatBarrier

\end{document}